\documentclass[letterpaper, 10 pt, conference]{ieeeconf}  

\IEEEoverridecommandlockouts                              
\usepackage{graphicx}
\usepackage{amsmath} 
\usepackage{amssymb}  
\usepackage{subcaption}
\usepackage{booktabs}
\usepackage{arydshln}

\usepackage[breaklinks=true,bookmarks=false,hidelinks]{hyperref}

\title{\LARGE \bf
Constitutional Precedent of Amicus Briefs*
}

\author{Allen Huang and Lars Roemheld$^{1}$
\thanks{*This is a final project for Sharad Goel's winter 2016 course MS\&E330 at Stanford University. We are grateful for an inspiring quarter.}
\thanks{$^{1}$Management Science \& Engineering, Stanford University, California. {\tt\small \{allenh, roemheld\}@stanford.edu}}
}

\begin{document}

\maketitle
\thispagestyle{empty}
\pagestyle{plain}

\begin{abstract}

We investigate shared language between U.S. Supreme Court majority opinions and interest groups' corresponding amicus briefs. Specifically, we evaluate whether language that originated in an amicus brief acquired legal precedent status by being cited in the Court's opinion. Using plagiarism detection software, automated querying of a large legal database, and manual analysis, we establish seven instances where interest group amici were able to formulate constitutional case law, setting binding legal precedent. We discuss several such instances for their implications in the Supreme Court's creation of case law. 

\end{abstract}

\section{INTRODUCTION}

Through amicus briefs, the U.S. Supreme Court allows third parties to contribute important facts, make legal arguments, or point to economic and policy implications of the case. Intended as a means for experts (friends of the court, \emph{amicus curiae}) to help the court make more informed decisions, it has been noted that the past decades have seen a significant increase in amicus briefs submitted by interest groups of all sorts~\cite{kearney}. The creation of legal briefs is costly, and it may be assumed that most amicus briefs present a group's vested interest, rather than neutral advice~\cite{collins}.

Amicus briefs are posted to the court at two stages of proceedings: first, during petition for certiorari, during which the Supreme Court decides whether an appeal to hear a case is granted (certiorari stage). Second, after certiorari has been granted and in preparation for a hearing (merits stage). Together with two other document types, the original petition for certiorari and the merits briefs (formal arguments by the parties directly involved in a case), amicus briefs provide the argumentative basis on which a case is then finally heard. After all proceedings, the court's final decision is documented in a written majority opinion~\cite{corley}. Individual justices may file minority opinions if they disagree in part or in whole with the majority, however only the majority opinion sets precedent: a written decision by the Supreme Court is binding federal law for all U.S. courts in that similar cases must follow the ruling.

In practice, there are very few restrictions on who is allowed to submit an amicus brief~\cite{collins}, and some briefs have been shown to be of spurious scientific value. In some instances, this has not stopped the court from relying on amicus briefs in its ruling~\cite{larsen}. In times of increasing complexity of information, and increasing ease of fabricated or misinterpreted data, the Supreme Court's responsibility of evaluating the quality of received briefs is becoming more and more challenging. This is especially true of amicus briefs that claim to be based on privileged, and hence unpublished, data (see~\cite{larsen} for examples).

We argue that the exact wording of Supreme Court opinions matters, regardless of factual accuracy: by setting precedent, each Supreme Court opinion \emph{creates} constitutional case law, which subsequent jurisdictions have to respect. When these jurisdictions, whether Supreme Court or lower courts, apply precedent opinions to their new cases, exact language matters, because it determines the realm of possible interpretation---in some cases, this may have unforeseen implications.

Indeed, some interest groups have openly stated that their motivation in regularly submitting amicus briefs to the Supreme Court is to over time ``mold'' cumulative case law to favor their positions~\cite{oconnor}. This is achieved, of course, by directly influencing Justice voting behavior, which has been studied widely \cite{boxsteffensmeier, caldeira, collins2007, collins2008, kearneymerill}.\footnote{%
Despite these studies, ``Supreme Court decisionmaking [is] a complex phenomenon,''~\cite{collins}, and getting robust quantitative estimates on the isolated effect of briefs on Justice votes is a tremendous challenge.}
It is also achieved, however, by shaping the specific wording of the majority opinion, which determines the precedent set for future interpretation.

Previous studies have compared opinion language with amicus language and found frequent instances of overlapping language; this can partly be attributed to the work of Supreme Court legal clerks, who in drafting a written opinion are tasked with summarizing the arguments that led to the majority's decision. In this process of collating arguments, it is common practice to rely on wording provided by the relevant documents and briefs (petition documents, merits briefs, amicus briefs), sometimes without explicit citation of sources~\cite{larsen, collins, corley, feldman2015, feldman2016}.

Building on this body of literature, this study asks if direct evidence can be found that interest groups acting as amici were allowed to formulate constitutional case law. More specifically, we aim to detect and quantify instances where amicus language has set precedent, after being used in the Supreme Court's majority opinion. 

Where it is possible to plausibly argue that wording by third-party interest groups was awarded binding constitutional precedent through opinion text, it is likely that ``molding'' has occurred to some degree. In some instances, this may be concerning, when partisan positions shape constitutional case law; in others, it may be the result of assiduous amicus work. In any case, it is revealing to trace precedent language back to its origins within the legal process, and to become conscious of the ``molding'' power granted to amici. We are unaware of any previous work to this end.

\section{DATA AND METHODS}

While all Supreme Court cases are publicly available, we could not find a comprehensive, readily available dataset. We therefore collected our own data, comprising almost all Supreme Court cases from October terms (OT) 2007 through 2011 (October 2007-June 2012).\footnote{All code used to obtain and analyze our data is available at \url{https://github.com/larsroemheld/mse330-scotus-briefs}.}

It is common practice for the Supreme Court to bundle several cases in one opinion; we therefore follow the literature in using dockets as the unit of our analysis. Overall, our data comprises 408 dockets, which we collected from the coverage on the website SCOTUSblog.\footnote{\url{http://www.scotusblog.com/case-files/terms/}} Due to lacking coverage, this is 12 dockets short of all dockets from the 2007-2011 October terms.

\begin{figure}[tb]
  \centering
  \includegraphics[width=0.9\linewidth]{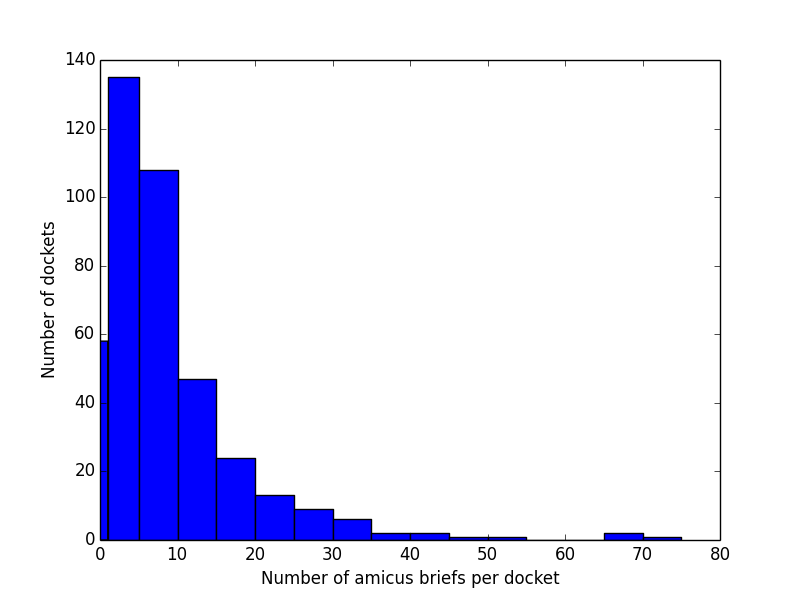}
  \caption{Histogram of the number of amicus briefs per Supreme Court docket, OT 2007-2011 (408 dockets in total, 3196 amicus briefs). Note that 57 dockets had no associated amicus briefs.}
  \label{briefshistogram}
\end{figure}

For each docket, we downloaded all case metadata, the majority opinion, and all amicus briefs. This led to a total of 3196 amicus briefs, the distribution of which is shown in figure~\ref{briefshistogram}. After crawling the page, we had to convert all documents from pdf format to plain text. This introduced some encoding noise, but worked remarkably well overall.

Following~\cite{feldman2016} and~\cite{corley}, we then used the open-source plagiarism\footnote{%
We note that the use of \emph{plagiarism}-detection software may be semantically misleading: we do not attempt to detect ``plagiarism'' of any kind. Even when language is clearly overlapping between an amicus brief and the majority opinion, and said language is not cited, the fact that the ``friend of the court'' wrote the brief with the expressed purpose of being used in the proceedings seems to mitigate the notion of ``plagiarism.'' Notwithstanding these considerations, it may be preferable practice to cite sources where they influence opinion wording~\cite{feldman2016}.
} checker WCopyfind\footnote{Lou Bloomfield: WCopyfind 4.1.4, \url{http://plagiarism.bloomfieldmedia.com/wordpress/software/wcopyfind/}} to compare each opinion with the docket's associated amicus briefs. We deviated from the default settings used in the literature, defining overlapping language as a sequence of 10 words or more, of which at least 80\% of words need to match exactly. Our increase of the minimum match-length from the default 6 words to 10 words was necessary to avoid common turns of phrases and other snippets, and focus on more meaningful language overlap.

We found 21,361 instances of overlapping language between majority opinions and their corresponding amicus briefs. Even after increasing the requirements of language overlap, the majority of matches are still common phrases or shared references to previous (case) laws. Within these raw language matches, we then searched instances where amicus language had successfully reached precedent status.

To do so, we used Lexis Advance\footnote{\url{https://advance.lexis.com/}}, a comprehensive, proprietary legal database, to establish whether a language snippet had appeared in any court case (including lower courts) before the Supreme Court case in question. If the snippet had appeared, we concluded that the language snippet could not have originated in the amicus brief: it was either a phrase commonly found in legal proceedings, or had some precedent status prior to the Supreme Court case in question. To focus only on snippets that achieved some form of precedent after being used in the majority opinion, we then queried whether the exact snippet appeared in later cases, in any court.

This process of querying the Lexis Advance database for all 21,361 matches yielded 944 language snippets that satisfied all of the following conditions:
\begin{enumerate}
\item{The snippet is at least 10 words long.}
\item{The snippet is used in an amicus brief.}
\item{The snippet is used in the corresponding majority opinion, with an overlap of at least 80\% of words.}
\item{The exact snippet from the opinion was never used in a court of law before the Supreme Court case.}
\item{The exact snippet from the opinion was used at least once in subsequent cases in any court of law.}
\end{enumerate}

These conditions make it likely that a language snippet did indeed originate in an amicus brief and achieved legal precedent through inclusion in a majority opinion. Furthermore, the precedent has already manifested itself at least once. Unfortunately, inspection showed that the exact-match queries performed on Lexis Advance were a limitation, as even very minor changes implied that either (i) previous use of the snippet was not detected, or (ii) later citation of the snippet as used in the opinion was not picked up. Since Lexis Advance does not allow queries for softer language overlap, this is a limitation we could not address automatically. Given the 944 snippets found by our process, we therefore started a manual analysis, tracing back the origins of snippets and evaluating relevance of the generated evidence.

Despite the limitations, we believe that ours is a previously unused method of tracing specific language through its propagation in legal proceedings, from a likely originator through citation and ``borrowed language'' into later cases and applications.

\section{RESULTS}

Given the stark limitations of exact-match queries on previous and following cases, our results are surprisingly rich. Due to time constraints, we only analyzed 100 out of the 944 snippets originally found; we found seven instances of precedent-setting amicus brief language that warrant further attention.

Manual analysis of our results is a challenging task, as legal language needs to be carefully considered for the following conditions. This decision of whether or not a snippet is an amicus-set precedent (ASP) is not always straight-forward.
\begin{enumerate}
\item The snippet is ``interesting,'' i.e. the language overlap is more significant than just a turn of phrase (noise).
\item The snippet did not have precedent before being used by the amicus. Specifically, it is not a citation or direct paraphrase from previous legislation or Supreme Court opinions (this check was made necessary by the limitation of exact-match queries on Lexis Advance).
\item The language did not originate from petition documents. When an amicus' paraphrase of language from the petition for certiorari appears in the majority opinion, this likely just indicates that the amicus understood the ``crux'' of the case itself.
\end{enumerate}

As we had to limit our analysis to 100 snippets, we looked at the 50 snippets with most appearances in later court cases (893--23 later appearances), and then looked at 50 randomly selected snippets from the remaining results (22--1 appearances) to get a more representative idea of our data.

Within the 50 snippets with most later appearances (i.e. cited the most), we found one ASP. In the 50 randomly selected snippets, we found 6 ASP -- if this random sample is representative, we would expect around 110 instances of ASP in our complete dataset. 

As expected, the rate of ASP was much higher outside of the most-appearances results. This is intuitive, as oft-cited snippets are more likely to be former precedents that our methods failed to pick up.


In the remainder of this section, we present a selection of ASP found using the methods described. 



\subsection{Third-party language}
First, in Wal-Mart v. Dukes (OT 2010), Wal-Mart petitioned against the largest class action law suit in U.S. history, in which six women held that the company systematically disadvantaged women employees. In a 5-4 decision for Wal-Mart, the court held that the six respondents could not represent all 1.5 million women employees of the company as one class, overruling previous decisions granting the class action lawsuit.

In its amicus brief in support of Wal-Mart, the Intel Corporation cites from a law review on class action lawsuits. While published in a respected law review, this article to this point had been one academic's work---after the Supreme Court joined Intel in citing verbatim from the article, its language gained federal precedent status. Since 2011, the exact wording has been cited in 108 cases (See Fig. \ref{tab:walmart-dukes}).

This is a relatively common type of ASP: a third party's language gains precedent after being used in an amicus brief. Careful consideration is required to parse out whether an amicus has indeed popularized a third party's wording, or whether the brief has merely understood the relevance of an outside source. We note that the mere fact that language originates from a non-precedent third party source does not make the overlap uninteresting, since the amicus can choose from a wide range of ``scientific'' opinions and wording to use.

Other examples of ASP produced from third party language follow. In Northwest Austin Municipal Utility District Number One v. Holder, we found that the amici were the ones to first introduce the \emph{The Promise and Pitfalls of the New Voting Rights Act}, an article in the Yale Law Journal, into case law. Similarly, in Florence v. Board of Chosen Freeholders of the County of Burlington, the amici were able to introduce the work of a study done by the New Jersey Commission of Investigation (\emph{Gangland Behind Bars-How and Why Organized Criminal Street Gangs Thrive in New Jersey's Prisons}). The study's wording has since re-appeared in proceedings twice (See Fig. \ref{tab:science-examples}). Another example, presented in Fig. \ref{tab:random-examples}, shows an amicus brief introducing language from an opinion of the Supreme Court of the Territory of Michigan -- which awarded federal precedent to state law language.

\subsection{Original amicus language}
A second class of overlap can be found in Caperton v. A.T. Massey Coal Co. Caperton challenged the constitutionality of a judge not recusing himself from a case in which one of the parties donated 3 million dollars to his election campaign. In a 5-4 decision, the Supreme Court decided that it was in fact unconstitutional for a judge to not recuse himself based on the Due Process clause of the 14th Amendment. 

The Conference of Chief Justices filed a brief in which they discuss the implications of the upcoming decision and how it might affect public trust of the judicial system. The majority opinion directly attributes \emph{original} language from the brief as a concise summarization of one of the key points of the case. In this case, original brief language achieved federal precedent. Indeed, the exact opinion quotation in Fig. \ref{tab:caperton-atmassey} was cited 9 times subsequent to the decision.

This type of language overlap is most striking -- amicus language is used in the majority opinion as an authority, and manifests precedent in subsequent jurisdiction. It is somewhat reassuring that this is the only example of original amicus language precedent we found so far, and in this case it originated from esteemed justices. Even so, it seems worthwhile to ask how much leeway the amicus had in wording their brief, and how their particular choice of wording influenced future case law.

\subsection{Amicus paraphrase}
The final type of ASP we found is a significant paraphrase by the amicus of existing precedent. One example is given in Padilla v. Kentucky, where Padilla argued that it is the duty of counsel to advise on the risk of deportation in a criminal case. In a 7-2 decision, the Supreme Court decided that a lawyer is indeed required, within the context of the 6th Amendment, to advise his client on the risks of deportation. 

Elena Kagan, then solicitor general (and current Supreme Court Justice), filed a brief arguing for a stricter definition of the roles and duties of council in the U.S. adversarial system. In her brief, Kagan leverages a previous Supreme Court opinion, but emphatically paraphrases it with her own language, which gets quoted directly by the opinion (see Fig. \ref{tab:padilla-kentucky}).

\begin{figure*}[htbp]
\begin{center}
\begin{tabular}{p{0.4\textwidth} : p{0.4\textwidth}}
``The `crux' of Rule 23(b)(2), however, is the `indivisible nature of the injunctive or declaratory remedy warranted--the \emph{notion that the conduct is such that it can be enjoined or declared unlawful only as to all of the class members or as to none of} them.' Richard A. Nagareda, Class Certification in the Age of Aggregate Proof. 84 N.Y.U. L. REV. 97, 132 (2009).'' &
``The key to the (b)(2) class is `the indivisible nature of the injunctive or declaratory remedy warratned--the \emph{notion that the conduct is such that it can be enjoined or declared unlawful only as to all of the class members or as to none of} them.' Nagareda, 84 N.Y.U. L. Rev., at 132.'''
\\
\setlength\topsep{0pt}
\begin{flushright}
{\small (Brief for the Intel Corporation in Support of Petitioner)}
\end{flushright}
&
\setlength\topsep{0pt}
\begin{flushright}
{\small (Opinion by Scalia, Jun 2011)}
\end{flushright}
\end{tabular}
\end{center}
\caption{Wal-Mart v. Dukes. This is one example of how an amicus brief first quoted a third party, whose wording later achieved precedent to the level of constitutional case law through the majority opinion.  The exact quotation (in italics) has since been subsequently cited verbatim in court 108 times.}
\label{tab:walmart-dukes}
\end{figure*}

\begin{figure*}[htbp]
\begin{center}
\begin{tabular}{p{0.4\textwidth} : p{0.4\textwidth}}
``As judicial election campaigns become costlier and more politicized, \emph{public confidence in the fairness and integrity of the nation's elected judges may be imperiled}.'
&
``The Conference of the Chief Justices has underscored that the codes are "[t]he principal safeguard against judicial campaign abuses" that threaten to imperil "\emph{public confidence in the fairness and integrity of the nation's elected judges}." Brief for Conference of Chief Justices as \emph{Amicus Curiae} 4, 11.''
\\
\setlength\topsep{0pt}
\begin{flushright}
{\small (Conference of Chief Justices in Support of Neither Party)}
\end{flushright}
&
\setlength\topsep{0pt}
\begin{flushright}
{\small (Opinion by Kennedy, Jun 2009)}
\end{flushright}
\end{tabular}
\end{center}
\caption{Caperton v. A.T. Massey Coal Company, Inc. This exemplifies \emph{original} amicus brief language being cited and used as an authority, and thereby achieving precedent. The exact quotation (in italics) has since been subsequently cited verbatim in court 9 times.}
\label{tab:caperton-atmassey}
\end{figure*}

\begin{figure*}[htbp]
\begin{center}
\begin{tabular}{p{0.4\textwidth} : p{0.4\textwidth}}
``Because counsel€™'s function is to make the adversarial
testing process work in the particular case,
Strickland, 466 U.S. at 690, \emph{counsel is not constitutionally
required to provide advice on matters that will not
be decided in the criminal case}: matters that have nothing
to do with the defendant'€™s guilt or innocence of the
charges and that are not part of the punishment that the
prosecution seeks to impose for the offense. '' &
``The Solicitor General has urged us to conclude that Strickland applies to Padilla's claim only to the extent that he has alleged affirmative misadvice. In the United States' view, \emph{"counsel is not constitutionally required to provide advice on matters that will not be decided in the criminal case} . . .," though counsel is required to provide accurate advice if she [*370] chooses to discusses these matters. Brief for United States as Amicus Curiae 10.''
\\
\setlength\topsep{0pt}
\begin{flushright}
{\small (Brief for the United States of America in Support of Affirmance)}
\end{flushright}
&
\setlength\topsep{0pt}
\begin{flushright}
{\small (Opinion by Stevens, Mar 2010)}
\end{flushright}
\end{tabular}
\end{center}
\caption{In Padilla v. Commonwealth of Kentucky, Elena Kagan acting as Solicitor General filed a brief in which she paraphrased existing federal case law, to argue her point. This paraphrasing has been used in the majority opinion, and the exact quotation (in italics) has since been subsequently cited verbatim in court 2 times.}
\label{tab:padilla-kentucky}
\end{figure*}
\begin{figure*}
\begin{center}
\begin{tabular}{p{0.02\textwidth} p{0.4\textwidth} | p{0.4\textwidth}}
(1)&
``Since Section 5 continues to burden only certain
jurisdictions, the reauthorization can only be justified
on evidence showing how differences in minority voter
discrimination exist between the covered and
noncovered jurisdictions. Nathaniel \emph{Persily, The
Promise and Pitfalls of the New Voting Rights Act,
117} Yale L.J. 174, 195 (2007).''
&
``Congress heard warnings from supporters of extending § 5 that the evidence in the record did not address "systematic differences between the covered and the non-covered areas of the United States[,] . . . and, in fact, the evidence that is in the record suggests that there is more similarity than difference." The Continuing Need for Section 5 Pre-Clearance: Hearing before the Senate Committee on the Judiciary, 109th Cong., 2d Sess., 10 (2006) (statement of Richard H. Pildes); see also \emph{Persily, The Promise and Pitfalls of the New Voting Rights Act, 117 Yale} L. J. 174, 208 (2007)''
\\
&
\setlength\topsep{0pt}
\begin{flushright}
{\small (Pacific Legal Foundation, the Center for Equal Opportunity)}
\end{flushright}
&
\setlength\topsep{0pt}
\begin{flushright}
{\small (Opinion by Roberts, Jun 2009)}
\end{flushright}
\\
(2)&
``In 2009, the New Jersey State Commission of Investication ("SCI") authored a study entitled \emph{Gangland Behind Bars}. The investigation revealed that nearly 150,000 documented members of criminal street gangs are currently incarcerated in federal, state and local correctional facilities around the nation.''
&
``Jails and prisons also face grave threats posed by the increasing number of gang members who go through the intake process. See Brief for Policemen's Benevolent Association, Local 249, et al. as \emph{Amici Curiae} 14 (hereinafter PBA Brief); New Jersey Comm'n of Investigation, \emph{Gangland Behind Bars: How and Why Organized Criminal Street Gangs Thrive in New Jersey's Prisons... And What Can Be Done About It}.''
\\
&
\setlength\topsep{0pt}
\begin{flushright}
{\small (Policemen's Benevolent Assoc. Local 249 et al.)}
\end{flushright}
&
\setlength\topsep{0pt}
\begin{flushright}
{\small (Opinion by Kennedy, Apr 2012)}
\end{flushright}
\end{tabular}
\end{center}
\caption{Further examples of how amici introduced studies and articles into mainstream case law. (Note that the amicus brief in (2) cited the entire source in the table of authorities; the shown citation is the context in which it was used.)}
\label{tab:science-examples}
\end{figure*}

\begin{figure*}
\begin{center}
\begin{tabular}{p{0.02\textwidth} p{0.4\textwidth} | p{0.4\textwidth}}
(1)&
``The Michigan Territory'€™s Supreme Court,
also in a libel case, explained that the Constitution
"\emph{grants to the citizen the right to keep and bear arms.
But the grant of this privilege cannot be construed
into a right in him who keeps a gun to destroy his}
neighbor." United States v. Sheldon, 5 Blume Sup.
Ct. Trans. 337, 1829 WL 3021, at *12.''
&
``An 1829 decision by the Supreme Court of Michigan said: "The constitution of the United States also grants \emph{to the citizen the right to keep and bear arms. But the grant of this privilege cannot be construed into the right in him who keeps a gun to destroy his} neighbor. No rights are intended to be granted by the constitution for an unlawful or unjustifiable purpose." United States v. Sheldon, in 5 Transactions of the Supreme Court of the Territory of Michigan 337, 346 (W. Blume ed. 1940)[...].''
\\
&
\setlength\topsep{0pt}
\begin{flushright}
{\small (Brief for the CATO Institute and History Professor Joyce Lee Malcolm in Support of Respondent)}
\end{flushright}
&
\setlength\topsep{0pt}
\begin{flushright}
{\small (Opinion by Scalia, Jun 2008)}
\end{flushright}
\\
(2)&
``recognizing residual tort liability for anyone `who intentionally causes injury to another,' including `where
a person \emph{defrauds another for the purpose of causing
pecuniary harm to a third} person'''
&
``And the Restatement specifically recognizes `a cause of action' in favor of the injured party where the defendant `\emph{defrauds another for the purpose of causing pecuniary harm to a third} person.'''
\\
&
\setlength\topsep{0pt}
\begin{flushright}
{\small (Brief for the United States as Amicus Curiae Supporting Respondents)}
\end{flushright}
&
\setlength\topsep{0pt}
\begin{flushright}
{\small (Opinion by Thomas, Jun 2008)}
\end{flushright}
\end{tabular}
\end{center}
\caption{Further examples of how amici quoted legal sources that did not have federal precedent at the time of writing, where the opinion elevated the used language to federal precedent.}
\label{tab:random-examples}
\end{figure*}

\section{LIMITATIONS}

Our results come with several limitations. First of all, neither of us is versed enough in legal specifics to determine the relevance and factual correctness of our results; the evidence seems strong enough to be concerned about the results presented herein, but much of the language overlap we found may be deemed common practice or uninteresting by a lawyer.

The biggest technical limitation in our method is the exact-match search within Lexis Advance. With a ``softer'' method of searching, we would have been able to produce potentially more, higher-quality results automatically, for example because we could have ignored instances were an amicus cites previous federal precedent with only minuscule alterations. Weeding out such cases manually was an arduous, and potentially error-prone task. With high likelihood, we also missed later appearances of language matches that would have indicated stronger precedent-setting (see the discussion in section 2 above). 

Furthermore, for our original argument we would not have needed to consider only language that has already been cited: whether or not opinion language is cited, it has precedent. Such precedent may manifest itself at any point in the future. In this paper, we used the criterion of later appearance as a way to weed out ``uninteresting'' language -- more sophisticated techniques would make this unnecessary.

Due to the manual nature of our analysis, we could only inspect 100 instances out of our 944 matches. At best, this sample size may provide a preview of what further investigation may surface -- our results are by no means comprehensive yet.

Finally, we only took into account Supreme Court cases from the October Terms 2007 through 2011. Since the effects of precedent may take decades to manifest, it would be interesting to repeat this analysis with more historic data---which would mandate a less manual analysis, as significantly more results would be expected.

\section{CONCLUSION}

In this paper, we presented a novel way to find the origins of precedent-setting legal language in Supreme Court amicus briefs. We then trace this language from the brief through the majority opinion into precedent and later court cases.

Given the authors' lack of legal experience, our results should be taken with some precaution; however, we hold that our results seem convincing enough to warrant further investigation. We believe that our results strongly suggest that in certain instances, amici have been able to define language that set federal precedent. Whether or not this is problematic in itself is a philosophic question; we believe that in any case, precedent set by private interest groups is damaging to the public \emph{legitimacy} of the institution of the Supreme Court (see~\cite{feldman2016}). Along with other arguments made in~\cite{larsen}, we believe that our results suggest scrutiny of the role of the ``friends of the court'' in Supreme Court proceedings, and that the practice of amici may deserve a revision of standards.

Finally, we note that the relative consistency and conformity of legal language allows relatively straight-forward tracing of language snippets. To our knowledge, this characteristic has not yet been fully exploited by existing academic work.

\addtolength{\textheight}{-12cm}   






\end{document}